\pdfoutput=1

\documentclass[11pt]{article}

\usepackage[]{ACL2023}

\usepackage{times}
\usepackage{latexsym}

\usepackage[T1]{fontenc}

\usepackage[utf8]{inputenc}

\usepackage{microtype}

\usepackage{inconsolata}

\usepackage[labelformat=simple]{subcaption}

\usepackage{graphicx}
\usepackage{amsmath}
\usepackage{verbatim}
\usepackage{algorithm}
\usepackage{algpseudocode}
\usepackage{amsmath}
\usepackage{amssymb}
\usepackage{multirow}
\usepackage{makecell}
\usepackage{booktabs}
\usepackage{stfloats}
\usepackage[misc]{ifsym}
\usepackage{arydshln}

\newcommand*\samethanks[1][\value{footnote}]{\footnotemark[#1]}

\definecolor{empathy}{RGB}{168,201,234}
\definecolor{elicitation}{RGB}{245,203,177}
\definecolor{negative}{RGB}{199,231,171}
\definecolor{positive}{RGB}{255,212,196}

\title{Facilitating Multi-turn Emotional Support Conversation with Positive Emotion Elicitation: A Reinforcement Learning Approach}

\author{
    Jinfeng Zhou\textsuperscript{\rm 1,2}\thanks{\ \ Work done during internship at the CoAI Group.}\ \ \thanks{\ \ Equal contribution.}\quad
    Zhuang Chen\textsuperscript{\rm 1}\samethanks{}\quad 
    Bo Wang\textsuperscript{\rm 2}\thanks{\ \ Corresponding author.}\quad 
    Minlie Huang\textsuperscript{\rm 1} \\ 
    \small \textsuperscript{\rm 1}The CoAI Group, DCST, Institute for Artificial Intelligence, State Key Lab of Intelligent Technology and Systems, \\
    \small \textsuperscript{\rm 1}Beijing National Research Center for Information Science and Technology, Tsinghua University, Beijing 100084, China \\
    \small \textsuperscript{\rm 2}College of Intelligence and Computing, Tianjin University, Tianjin, China \\
    \small \texttt{jfzhou.mail@gmail.com, zhchen-nlp@mail.tsinghua.edu.cn, bo\_wang@tju.edu.cn, aihuang@tsinghua.edu.cn}
}

\begin{document}
\maketitle
\begin{abstract}

Emotional support conversation (ESC) aims to provide emotional support (ES) to improve one's mental state. Existing works stay at fitting grounded responses and responding strategies (e.g., \textit{question}), which ignore the effect on ES and lack explicit goals to guide emotional positive transition. To this end, we introduce a new paradigm to formalize multi-turn ESC as a process of positive emotion elicitation. Addressing this task requires finely adjusting the elicitation intensity in ES as the conversation progresses while maintaining conversational goals like coherence. In this paper, we propose \textsc{Supporter}, a mixture-of-expert-based reinforcement learning model, and well design ES and dialogue coherence rewards to guide policy's learning for responding. Experiments verify the superiority of \textsc{Supporter} in achieving positive emotion elicitation during responding while maintaining conversational goals including coherence.

\end{abstract}

\section{Introduction}

Emotional support (ES) aims to reassure a person to recover from emotional distress and improve one's mental state \cite{burleson2003emotional}. It is a manifestation of emotional intelligence in social interactions \cite{heaney2008social, atoum2018emotional}. Endowing ES into social dialogue systems for building helpful and trustful agents is an emerging trend \cite{DBLP:journals/tois/HuangZG20, rains2020support}. 

To achieve this goal, a typical practice is modeling empathy, which aims to perceive and understand the situation and feelings of others \cite{keskin2014isn}. 
Yet, the empathetic conversation \cite{DBLP:conf/acl/RashkinSLB19} is inherently deficient in providing ES as
(1) Lack of consideration of multi-turn conversation. Just making empathetic responses in each single dialogue turn leads to ignoring the user's feedback and mental state changes in multi-turn interaction. (2) Lack of awareness of emotional elicitation. Only emanating emotional resonance fails to help users jump out of negative mental states. Although \citet{DBLP:conf/acl/LiuZDSLYJH20} design emotional support conversation (ESC) task promising to remedy these deficiencies, existing works \cite{DBLP:conf/acl/TuLC0W022, DBLP:journals/corr/abs-2210-04242, DBLP:conf/ijcai/00080XXSL22} stay at fitting grounded responses and responding strategies (e.g., \textit{question}) while ignoring the effects of such efforts on ES. They do not fully model the essential working mechanism of ESC and lack explicit goals to guide a user's emotion to a positive transition in the multi-turn process. Thus, they are still insufficient to lay out an entire ESC process and cannot effectively improve one's mental state.

\begin{figure}[t]
\centering
\includegraphics[width=.98\columnwidth]{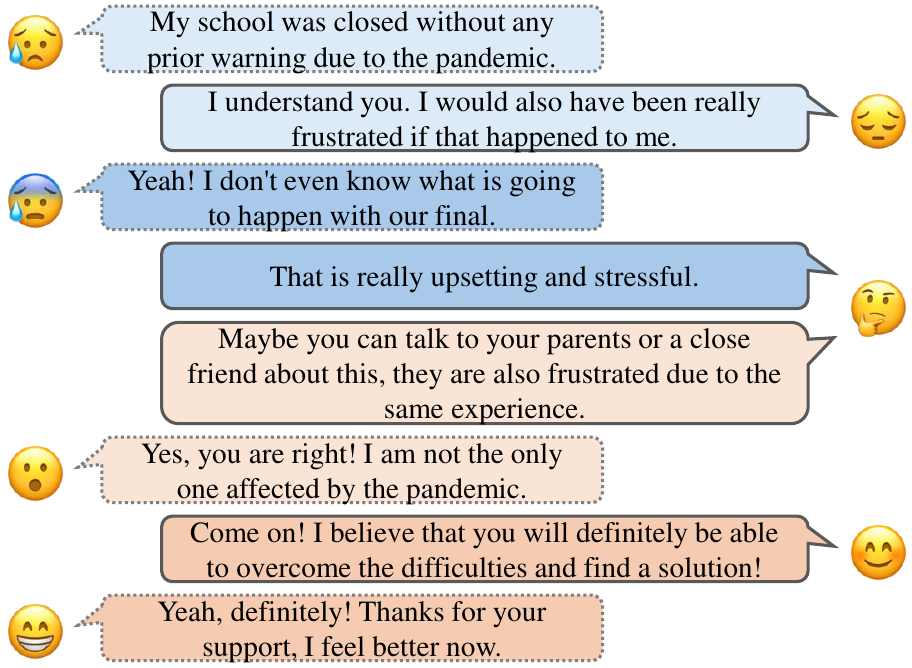} 
\caption{A simplified multi-turn ESC example between the user (\textit{left}) and agent (\textit{right}). The agent progressively adjusts the intensity of \colorbox{empathy}{\textit{empathy}} and \colorbox{elicitation}{\textit{elicitation}} to achieve the goal of improving the user's mental state.}
\label{introduction}
\end{figure}

To this end, we introduce multi-turn ESC with positive emotion elicitation, a new paradigm aims to progressively empathize and elicit users to reach a better mental state through multi-turn conversation. Addressing this task is challenging (an example is in Figure \ref{introduction}):
\textbf{First}, in a realistic multi-turn ESC, the user's emotions often transit towards positive (e.g., the user's emotion starts with negative and ends with positive, i.e., “\textit{My school was closed}” $\rightarrow$ “\textit{I feel better now}”) with fluctuation (e.g., the user's negative emotions in the first two turns gradually deepen, i.e., “\textit{My school was closed}” $\rightarrow$ “\textit{I don’t even know}”), which requires the agent to equip with the mechanism dealing with complex situations to respond satisfactorily \cite{DBLP:conf/iwsds/ShibataEK14, DBLP:journals/csl/YoshinoK15}.
\textbf{Second}, for ES, the ES response requires a delicate balance between empathy and elicitation. Only empathizing without eliciting falls into a negative emotional cycle, while the opposite setting brings a sense of distance in communication. 
They need to be progressively and purposefully adjusted in ongoing interactions, e.g., the agent expresses empathy of varying emotional polarity (\textit{negative} $\rightarrow$ \textit{negative} $\rightarrow$ \textit{positive}) and carefully increase the intensity of elicitation (\textit{only empathy} $\rightarrow$ \textit{weak elicitation} $\rightarrow$ \textit{strong elicitation}).
\textbf{Third}, for language expression, the ES response purposefully elicits positive emotions but should not undermine general conversational goals like coherence. Making an eliciting response that is out of the dialogue context, e.g., replacing “\textit{I understand you. I would ... happened to me.}” with “\textit{Come on! I believe ... find a solution!}”, may cause users to resent and block useful feedback.

In this paper, we propose \textbf{\textsc{Supporter}}\footnote{The project repository is available at \url{https://github.com/jfzhouyoo/Supporter}} to facilitate multi-turn emotional \textbf{\textsc{Support}} conversation with positive emotion \textbf{\textsc{e}}licitation using a mixture-of-expert(MoE) based \textbf{\textsc{r}}einforcement learning(RL).
MoE designs heuristic experts associated with specific tasks to learn diverse semantics by characterizing dialogue context, where: 
(1) To cope with the user's emotional fluctuation in the ongoing conversation, experts are devised as positive and negative experts as a whole;
(2) To inspire ES of responding, the emotion experts of MoE are designed to predict the user's emotional states that are possibly transited to;
(3) To inspire the expression of responding, the keyword experts of MoE are designed to predict the keywords that maintain the dialogue coherence.
With experts as candidates, our RL agent learns conversational semantic encoding policy and purposefully selects experts with expert selection policy for response generation.
To achieve the goal of positive emotion elicitation during responding while maintaining conversational goals like coherence, we optimize policy by carefully constructing the rewards:
(1) ES rewards consider the conversation progress to dynamically adjust the elicitation intensity of positive emotion;
(2) Dialogue coherence rewards involve keyword-level and sentence-level guides to finely maintain coherence.

Our contributions are summarized as follows:

(1) We introduce a new paradigm by carefully dissecting the challenges of formalizing multi-turn ESC as a process of positive emotion elicitation.

(2) We propose \textsc{Supporter}, an MoE-based RL model with carefully constructed ES and dialogue coherence rewards, elicits positive emotion during responding while maintaining dialogue coherence.

(3) Extensive experiments show the superiority of \textsc{Supporter} with automatic, interactive human, and novel ES and dialogue coherence evaluations.

\section{Related Work}

\paragraph{Empathetic Conversation}
To construct a warm dialogue system, a milestone is to endow it with empathy \cite{DBLP:conf/acl/RashkinSLB19}.
Considering affective empathy \cite{DBLP:conf/emnlp/LinMSXF19, DBLP:conf/emnlp/MajumderHPLGGMP20, DBLP:conf/coling/LiCRRTC20, DBLP:conf/aaai/LiLRRC22}, i.e., perceiving the user's emotion, and cognitive empathy \cite{DBLP:conf/acl/ZhengLCLH21, DBLP:conf/aaai/SabourZH22, DBLP:journals/corr/abs-2208-08845}, i.e., understanding the user's situation, puts the psychological theory of empathy into practice. 
Limited by focusing on a single-turn empathy and lack of emotional induction, it is difficult to achieve the higher goal of improving the user's mental state due to failure to help one jump out of the negative situation.

\paragraph{Emotional Support Conversation}
To remedy above deficiencies, \citet{DBLP:conf/acl/LiuZDSLYJH20} design ESC for providing ES in interactions. Our work is related to existing works on ESC but differs in task definition as we focus on enhancing the elicitation effect of positive emotion of responses instead of responding strategy prediction (e.g., \textit{question}) and grounded response generation. Although fusing knowledge \cite{DBLP:conf/acl/TuLC0W022, DBLP:conf/ijcai/00080XXSL22} and planning strategy \cite{DBLP:journals/corr/abs-2210-04242} are beneficial for word-overlap metrics (e.g., \textit{Bleu}), we argue whether the gains serve to ES is opaque and less convincing due to lacking corresponding evaluation mechanisms.

\paragraph{Positive Emotion Elicitation Conversation}
To free users from emotional distress and advance the conversation towards an optimistic state, positive emotion elicitation is an intuitive solution \cite{mishara2007helper, DBLP:journals/corr/abs-2105-08251}. Previous works \cite{DBLP:conf/acl/HasegawaK0T13, DBLP:conf/aaai/LubisSYN18, lubis2019dialogue, DBLP:journals/taslp/LubisSYN19} posit the emotional elicitation process as an ideal single-turn dialogue with linear emotional changes \cite{DBLP:journals/corr/abs-2204-10509}. However, realistic scenarios often involve multi-turn interactions with complex emotional fluctuations. To weaken the previous strong hypothesis, we extend positive emotion elicitation to ESC by well defining challenges, and take it as a real-world application of the solution.

\begin{figure*}[t]
\centering
\includegraphics[width=0.92\textwidth]{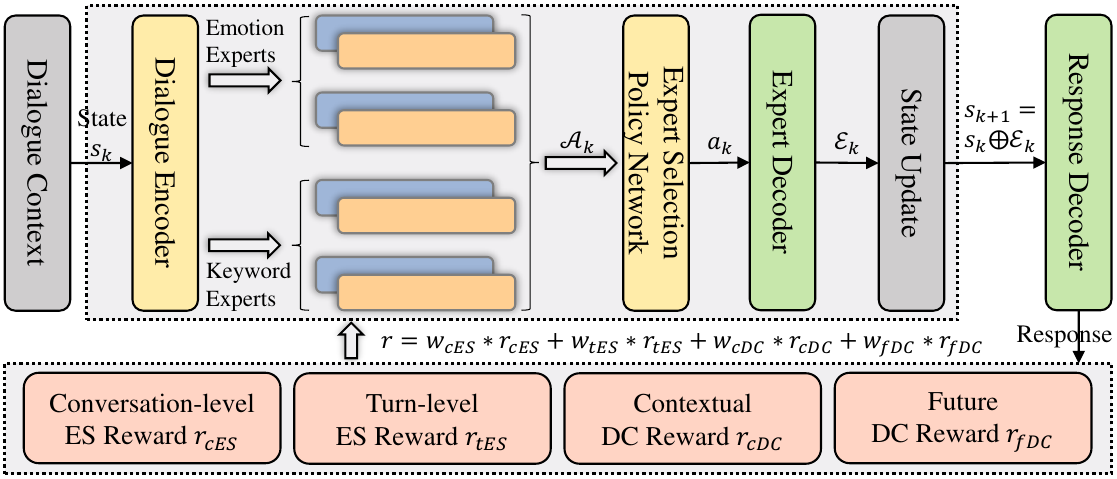} 
\caption{The architecture of the proposed \textsc{Supporter} model. \textit{DC} is an abbreviation for \textit{Dialogue Coherence}.}
\label{framework}
\end{figure*}

\section{Preliminaries}

At the $t$-th turn of dialogue, given dialogue context $C_t=\left\{x_1, y_1, \ldots, x_{t-1}, y_{t-1}, x_t\right\}$, our goal is to generate the response $y_{t}$ which serves to improve the user's mental state. To equip this ability, the response generation process should achieve specific goals related to ES and language expression.

\paragraph{ES for Positive Emotion Elicitation}
Providing effective elicitation during multi-turn ESC suffers from two issues: First, the elicitation intensity of positive emotion needs to be adjusted progressively as the conversation progresses. Maintaining weak elicitation (e.g., “\textit{I understand you}”) or strong elicitation (e.g., “\textit{Come on}”) may fail to shake one’s mental state. Second, the elicitation effect of positive emotion needs to be indirectly verified by the feedback from the user's next turn utterance. It means the elicitation intensity should consider the future fluctuation of the user’s emotional states. In this work, we construct conversation-level and turn-level ES rewards to guide the model's learning of elicitation policy and conduct corresponding automatic and interactive human evaluations for measuring the ES performance of responding.

\paragraph{Language Expression for Dialogue Coherence}
The purpose of generative processes to enhance elicitation induces two attendant issues: First, without proper controls may lead to greedily pursuing the goals of elicitation while discarding the contextual coherence, e.g., “\textit{Come on!}” with strong elicitation as a response in the context of the user continuing to express negative emotions. Second, whether the response meets the user's expectations needs feedback from the user's future utterance. It means maintaining coherence with future dialogue is also crucial. In this work, we construct contextual and future dialogue coherence rewards to guide the model's learning of bi-coherent expressions and perform the automatic and interactive human evaluation of conversational goals including coherence.

\section{Methodology}

In Figure \ref{framework}, our \textsc{Supporter} takes dialogue context as input to construct state sequence, which is encoded by a dialogue encoder as the conversational semantic encoding policy. The mixture-of-expert associated with emotion and keyword prediction tasks characterize state semantics to yield action candidates of the expert selection policy, which are purposefully selected for inducing state update. We use the updated state to generate response and further optimize the policy by measuring how well the response reaches the goal of ES and dialogue coherence with the well-designed parallel rewards.

\subsection{Multi-task Mixture-of-Expert}

As a key component of \textsc{Supporter}, we first introduce the structure of multi-task mixture-of-expert.

\paragraph{Dialogue Encoder}

Following \citet{DBLP:conf/acl/LiuZDSLYJH20}, the dialogue encoder is implemented with BlenderBot \cite{DBLP:conf/eacl/RollerDGJWLXOSB21}. Given an input sequence $X$, we concatenate all input tokens and prepend with a $[CLS]$ token, e.g., for the dialogue context, getting $[CLS] \oplus x_1 \oplus y_1 \ldots \oplus x_{t-1}$. The sequence is fed into the dialogue encoder to obtain the hidden state $\boldsymbol{H}_X$. We denote the sequence representation derived from $[CLS]$ as $\boldsymbol{h}_X$.

\paragraph{Emotion Experts}

To track possible transitions of user’s emotional states, emotion experts are associated with contextual and future user emotion predictions.
We extract $M$ fine-grained emotional reactions for each utterance in the corpus, which are inferred from COMET \cite{DBLP:conf/acl/BosselutRSMCC19} using the “\textit{xReact}” relation. Since emotional reactions are often emotional words (e.g., \textit{happy}, \textit{sad}), we use VAD \cite{DBLP:conf/acl/Mohammad18} to identify the emotional polarity of each word according to its valence as a positive or negative emotional category. The high-frequency categories are finally retained as supervised labels for the emotion prediction task.

We divide contextual emotion experts into positive and negative emotion experts, which are two MLP transforming $\boldsymbol{H}_X$ into $\boldsymbol{H}_{X,pos}$ and $\boldsymbol{H}_{X,neg}$:
\begin{equation}
\begin{aligned}
& \boldsymbol{H}_{X, p o s}=M L P_{p o s}\left(\boldsymbol{H}_X\right), \\
& \boldsymbol{H}_{X, n e g}=M L P_{n e g}\left(\boldsymbol{H}_X\right).
\end{aligned}
\end{equation}
We project the $[CLS]$ representations $\boldsymbol{h}_{X,pos}$ and $\boldsymbol{h}_{X,neg}$ of positive and negative experts to predict positive and negative emotion, respectively:
\begin{equation}
\begin{aligned}
P_{p o s} & =\operatorname{softmax}\left(\boldsymbol{W}_{p o s} \boldsymbol{h}_{X, p o s}\right), \\
P_{n e g} & =\operatorname{softmax}\left(\boldsymbol{W}_{n e g} \boldsymbol{h}_{X, n e g}\right),
\end{aligned}
\end{equation}
which is supervised by the positive and negative emotions collected in the $e_{pos}^*$ and $e_{neg}^*$ sets of the user's last utterance in the dialogue context using cross-entropy loss:
\begin{equation}
\begin{aligned}
& L_{p o s}^{c t x-e m o}=-\frac{1}{\left|e_{p o s}^*\right|} \sum_{i=1}^{\left|e_{p o s}^*\right|} \log P_{p o s}\left(e_i^*\right), \\
& L_{n e g}^{c t x-e m o}=-\frac{1}{\left|e_{n e g}^*\right|} \sum_{i=1}^{\left|e_{n e g}^*\right|} \log P_{n e g}\left(e_i^*\right). \\
\end{aligned}
\end{equation}
Note that an utterance may be inferred to the emotions with different polarities due to cognitive differences \cite{westbrook2011introduction, DBLP:journals/corr/abs-2208-08845}. For future emotion experts, we adopt the above method to get $L_{pos}^{ftr-emo}$ and $L_{neg}^{ftr-emo}$ losses and train them to predict the positive and negative emotions of the user's future utterance (i.e., next turn utterance). In this way, emotion experts can learn various emotion-level features by $L_{emo}$ loss: $L_{emo}=L_{pos}^{ctx-emo}+L_{neg}^{ctx-emo}+L_{pos}^{ftr-emo}+L_{neg}^{ftr-emo}$.

\paragraph{Keyword Experts}

To meet the need for dialogue coherence, keyword experts are associated with keyword predictions that act on maintaining coherence with contextual and future utterances.
Here, a bidirectional emotion keyword graph $\mathcal{G}$ is constructed, which is also used in coherence rewards designing (a construction example is in Appendix \ref{sec:graph}).
We extract the salient keywords of each utterance in the corpus as vertices using a rule-based approach \cite{DBLP:conf/acl/TangZXLXH19}, and employ VAD to identify the emotional polarity of each keyword. The pointwise mutual information (PMI) \cite{DBLP:conf/acl/ChurchH89} is adopted to construct bidirectional edges by characterizing the association between keyword pairs, where the \textit{forward} edge depicts the keyword pairs extracted from the context and response, and the \textit{backward} edge depicts the ones are from the future utterance and response. We further construct \textit{positive} edges to describe the keywords with positive tail vertices, and \textit{negative} edges are negative ones. Finally, each head vertex selects the tail vertices with the top PMI scores for building connections. The vertices of $\mathcal{G}$ serve as supervised labels for the keyword prediction task.

Contextual keyword experts are transformed similarly to emotion experts, and their $[CLS]$ representations $\boldsymbol{h}_{X,pos}^{ctx-kws}$ and $\boldsymbol{h}_{X,neg}^{ctx-kws}$ can be obtained from positive and negative keyword experts $\boldsymbol{H}_{X,pos}^{ctx-kws}$ and $\boldsymbol{H}_{X,neg}^{ctx-kws}$, respectively. We infer the one-hop neighbors of contextual keywords from the “\textit{forward-positive}” and “\textit{forward-negative}” relations respectively in $\mathcal{G}$ to enhance the perception of the target keywords in the golden response. Specifically, we use attention \cite{DBLP:journals/corr/BahdanauCB14} to obtain fused embeddings $\boldsymbol{e}_{pos}^{ctx-kws}$ and $\boldsymbol{e}_{neg}^{ctx-kws}$:
\begin{equation}
\begin{aligned}
& \boldsymbol{e}_{pos}^{c t x-k w s}=\operatorname{Attention}(\boldsymbol{h}_{X, p o s}^{c t x-k w s}, \boldsymbol{E}_{pos}^{c t x-k w s}), \\
& \boldsymbol{e}_{neg}^{c t x-k w s}=\operatorname{Attention}(\boldsymbol{h}_{X, n e g}^{c t x-k w s}, \boldsymbol{E}_{neg}^{c t x-k w s}),
\end{aligned}
\end{equation}
where $\boldsymbol{E}_{pos}^{ctx-kws}$ and $\boldsymbol{E}_{neg}^{ctx-kws}$ are positive and negative neighbor embedding matrices that share parameters with the dialogue encoder. We then concatenate $\boldsymbol{e}_{pos}^{ctx-kws}$ and $\boldsymbol{e}_{neg}^{ctx-kws}$ with $\boldsymbol{H}_{X,pos}^{ctx-kws}$ and $\boldsymbol{H}_{X,neg}^{ctx-kws}$ respectively at the token level, and use an MLP layer to fuse them to obtain keyword-enhanced experts $\boldsymbol{H}_{X,pos-kws}^{ctx-kws}$ and $\boldsymbol{H}_{X,neg-kws}^{ctx-kws}$:
\begin{equation}
\begin{aligned}
& \boldsymbol{H}_{X, pos-kws}^{ctx-kws}[i]=\operatorname{MLP}(\boldsymbol{H}_{X, pos}^{ctx-kws}[i] \oplus \boldsymbol{e}_{pos}^{ctx-kws}) \\
& \boldsymbol{H}_{X, neg-kws}^{ctx-kws}[i]=\operatorname{MLP}(\boldsymbol{H}_{X, neg}^{ctx-kws}[i] \oplus \boldsymbol{e}_{neg}^{ctx-kws})
\end{aligned}
\end{equation}
Further, we take the positive and negative keywords in the golden response as supervision to optimize the $L_{pos}^{ctx-kws}$ and $L_{neg}^{ctx-kws}$ losses adopting cross-entropy (this process can refer to above emotion prediction task). Similarly, multi-hop reasoning on $\mathcal{G}$, i.e., “\textit{forward} $\rightarrow$ \textit{forward} $\rightarrow$ \textit{backward-positive}” and “\textit{forward} $\rightarrow$ \textit{forward} $\rightarrow$ \textit{backward-negative}” (clarified in Appendix \ref{sec:graph}), is performed to obtain keywords coherent with the future utterance. Taking the positive and negative keywords in future utterance as the prediction target, the keyword-enhanced future keyword experts can be optimized by $L_{pos}^{ftr-kws}$ and $L_{neg}^{ftr-kws}$ losses. In this way, keyword experts can learn various expression-level features by $L_{kws}$ loss: $L_{kws}=L_{pos}^{ctx-kws}+L_{neg}^{ctx-kws}+L_{pos}^{ftr-kws}+L_{neg}^{ftr-kws}$.

\paragraph{Multi-task Training}

To make the experts retain the primitive semantics without hindering their respective diversity, we give them a minor constraint. Specifically, we average the representations of emotion and keyword experts to get $\boldsymbol{h}_{X,exp}$, and make it close to sequence representation $\boldsymbol{h}_{X}$ by optimizing the MSE loss with a minor hyperparameter $\alpha$:
\begin{equation}
L_{mse}=\frac{\alpha}{d_h} \sum_{i=1}^{d_h}\left(\boldsymbol{h}_{X}[i]-\boldsymbol{h}_{X, exp}[i]\right)^2,
\end{equation}
where $d_{h}$ is the dimension of $\boldsymbol{h}_{X}$. Then, we jointly train the multi-task MoE by optimizing $L_{exp}$ loss: 
\begin{equation}
L_{exp}=L_{emo}+L_{kws}+L_{mse}.
\end{equation}

\subsection{MoE-based Reinforcement Learning}

We use the standard reinforcement learning framework \cite{sutton2018reinforcement} as the backbone.

\paragraph{State}

We concatenate the dialogue context and the extracted keywords as the initial state $s_1 \in \mathcal{S}$, i.e., $s_{1}=\{C,C_{kws}\}$ (we omit the subscript $t$ of dialogue context $C_{t}$ for simplicity). At each step, the prompt token sequence $\mathcal{E}$ generated by the policy determined expert (i.e., action) triggers an update of the state. We record the observed state $s_k \in \mathcal{S}$ at $k$-th step, i.e., $s_k=\left\{C, \mathcal{E}_1, \ldots, \mathcal{E}_{k-1}\right\}$, which is encoded by the dialogue encoder to get $\boldsymbol{H}_{S,k}$ and $\boldsymbol{h}_{S,k}$. We concatenate sequence representations of historical states to obtain current state embedding $\boldsymbol{s}_{k}=\boldsymbol{h}_{S, 1} \oplus \ldots \oplus \boldsymbol{h}_{S, k}$. If $k$ is smaller than the set maximum iteration steps $K$, we pad $\boldsymbol{s}_{k}$ with zeros for fixing dimension. Note that when $k>1$, we discard the keywords $C_{kws}$ because: (1) It has already acted on the first iteration; (2) The input sequence length is limited due to the constraint of the pre-trained model (i.e., BlenderBot).

\paragraph{Action}

The action space $\mathcal{A}_{k}$ at $k$-th step is defined as the multi-task associated experts transformed by state $s_{k}$. At state $s_{k}$, our agent learns to choose an expert in $\mathcal{A}_{k}$ as expert action $a_{k}$. We utilize a BlenderBot-based dialogue decoder to generate expert prompt $\mathcal{E}_{k}$ of $a_{k}$.

\paragraph{Policy}

Besides the above dialogue encoder as the semantic encoding policy network, we design an expert selection policy network using REINFORCE with baseline \cite{sutton2018reinforcement} that includes an actor network and a value network. Actor learns an expert finding policy $\pi_{\varphi}\left(a_k, s_k, \mathcal{A}_k\right)$ which selects the appropriate expert action $a_{k}$ based on the current state $s_{k}$ and action space $\mathcal{A}_{k}$ by emitting the probability distribution of actions in $\mathcal{A}_{k}$. The value network measures the value $Q_\delta\left(s_k\right)$ of state $s_{k}$ as the baseline in REINFORCE. Their network structures are defined as:
\begin{equation}
\begin{gathered}
\boldsymbol{o}_k=\eta\left(\left(\eta\left(\boldsymbol{s}_k \boldsymbol{W}_1\right) \boldsymbol{W}_2\right)\right), \\
\pi_{\varphi}\left(a_k, s_k, \mathcal{A}_k\right)=\phi\left(\boldsymbol{A}_k \odot \boldsymbol{o}_k \boldsymbol{W}_{\varphi}\right), \\
Q_\delta\left(s_k\right)=\boldsymbol{o}_k \boldsymbol{W}_\delta,
\end{gathered}
\end{equation}
where $\eta(\cdot)$ is an ELU activation function with a dropout layer, $\odot$ is the hadamard product, $\phi(\cdot)$ is the softmax function. $\boldsymbol{A}_{k}$ is a binarized vector for pruning the action space, and we set it as a full-one vector due to the small number of experts.

\paragraph{Rewards}

To guide policy learning, we reward the decision made at each step by measuring how well the response generated from updated state $s_{k+1}$ provides ES and maintains dialogue coherence.

(1) Conversation-level ES Reward: aims to dynamically adjust the elicitation intensity of positive emotion as the conversation progresses defined as:
\begin{equation}
\begin{gathered}
P E D_{c E S}=f_{E S}(y)-f_{E S}\left(c_t\right), \\
r_{c E S}=\sum_{t=1}^T \cos (\frac{\pi}{2} \cdot \frac{t}{M T}) \cdot P E D_{c E S}.
\end{gathered}
\end{equation}
Here, $f_{ES}(\cdot)$ measures the positive emotion level of an utterance using the emotion classification model developed by \citet{hartmann2022emotionenglish}. The model is trained on six datasets containing diverse text types and achieves 66\% accuracy for emotion classification. Positive emotion scores are collected as positive level. We encourage the positive emotion distance $PED_{cES}$ of the generated response $y$ and the contextual user's post $c_{t}$: (a) is non-negative, i.e., expressing empathy (equal to 0) or elicitation (greater than 0) is the underlying requirement; (b) synchronously increases with the dialogue turn $t$, i.e., the early stage of the conversation is dominated by empathy, and the latter is elicitation. $MT$ is the maximum turn of conversation, $T$ is current turn.

(2) Turn-level ES Reward: aims to capture the feedback of user's next turn emotion defined as:
\begin{equation}
\begin{gathered}
P E D_{t E S}=\left|f_{E S}(y)-f_{E S}\left(c_f\right)\right|, \\
r_{t E S}=\cos (\frac{\pi}{2} \cdot \frac{T}{M T}) \cdot \cos (\frac{\pi}{2} \cdot P E D_{t E S}).
\end{gathered}
\end{equation}
Here, $PED_{tES}$ measures the relative positive emotion distance between the generated response $y$ and the user's future (i.e., next turn) utterance $c_{f}$. We encourage $PED_{tES}$ to get smaller with the approaching of current turn $T$ to $MT$, i.e., supervising smooth elicitation in the latter stage and improving tolerance to emotional fluctuations.

(3) Contextual Dialogue Coherence Reward: aims to constrain generated response $y$ to maintain coherence with context $C$ by measuring their coherence at keyword-level and sentence-level. First, we reconstruct a dataset \cite{DBLP:conf/acl/LiuZDSLYJH20} containing coherent and incoherent context-response pairs, where the response of the incoherent pairs is an utterance randomly sampled from the dataset. Next, a BERT-based \cite{DBLP:conf/naacl/DevlinCLT19} text classification model $f_{cDC}$ is trained by feeding sentence-keyword pairs and achieves 85\% accuracy. We take the coherence probability as the coherence score, the reward is defined as:
\begin{equation}
r_{cDC}=f_{cDC}\left(C \oplus C_{kws}, y \oplus y_{kws}\right) \cdot e^{\frac{N_{c, kws}}{\left|y_{kws}\right|}-1},
\end{equation}
where $y_{kws}$ is the keyword set of $y$ and $N_{c,kws}$ is the number of keywords in $y_{kws}$ that are the \textit{forward} neighbors of contextual keywords in $\mathcal{G}$.

(4) Future Dialogue Coherence Reward: aims to introduce the consideration of coherence with the user's future utterance $c_{f}$. Similarly, we reconstruct a dataset \cite{DBLP:conf/acl/LiuZDSLYJH20} containing coherent and incoherent future utterance-response pairs and train another text classification model $f_{fDC}$ which achieves 77\% accuracy. The reward is defined as:
\begin{equation}
r_{fDC}=f_{fDC}\left(c_f \oplus c_{f_{kws}}, y \oplus y_{kws}\right) \cdot e^{\frac{N_{f, kws}}{\left|y_{kws}\right|}-1},
\end{equation}
where $N_{f,kws}$ is the number of keywords in $y_{kws}$ that have a \textit{backward} relation with keywords $c_{f_{kws}}$ of $c_{f}$ in $\mathcal{G}$.

(5) Total reward. The total reward is $r=w_{cES}*r_{cES}+w_{tES}*r_{tES}+w_{cDC}*r_{cDC}+w_{fDC}*r_{fDC}$.

\subsection{Optimization}

We set $K$-step iterations, and the goal of agent learning is to maximize the expected cumulative reward: $J_\theta=\mathbb{E}_\pi\left[\sum_{k=1}^K \gamma^k r_{k+1}\right]$, where $\theta$ is the learned parameter and $\gamma$ is the discount coefficient. The agent is optimized by $L_{agent}$ loss and its policy gradient is defined as:
\begin{equation}
\begin{aligned}
\nabla_\theta J_\theta=\mathbb{E}_\pi[\nabla_\theta \log \pi_{\varphi}(a_k, s_k, \mathcal{A}_k)(G-Q_\delta(s_k))],    
\end{aligned}
\end{equation}
where $G$ is the discounted cumulative reward from the initial state to the terminal state. Finally, we take the hidden state $\boldsymbol{H}_{S,K+1}$ of the state $s_{K+1}$ to generate the response, where the decoder is optimized by $L_{gen}$ loss:
\begin{equation}
L_{gen}=-\sum_{m=1}^M \log P(y_m \mid \boldsymbol{H}_{S, K+1}, y_{<m}).
\end{equation}

\paragraph{Warm Start}
We use the pretrained small version of BenderBot for initializing our model. The initial state is used as input to fine-tune the model for warm start by optimizing $L_{warm}=L_{exp}+L_{gen}$.

\paragraph{Joint Training}
Our model is finally jointly trained by optimizing $L_{joint}$ loss:
\begin{equation}
L_{joint}=L_{agent}+L_{gen}+\frac{1}{K+1} \sum_{k=1}^{K+1} L_{exp, k}
\end{equation}

\section{Experiments}

\begin{table}[t]
\centering
\resizebox{.95\columnwidth}{!}{
\begin{tabular}{c c c}
\hline
    \multirow{5}{*}{\makecell[c]{Corpus \\ Info.}} 
    & \#Dialogues & 1,053 \\
    & \#Utterances & 31,410 \\
    & Avg. length of dialogues & 29.8 \\
    & Avg. length of utterances & 17.8 \\
    & \#Split Ratio & 8:1:1 \\
\hline
    \multirow{5}{*}{\makecell[c]{Graph $\mathcal{G}$ \\ Info.}} 
    & \#Keywords & 2,433 \\
    & Avg. forward neighbors & 21.24 \\
    & Avg. backward neighbors & 21.17 \\
    & Avg. positive neighbors & 33.94 \\
    & Avg. negative neighbors & 8.46 \\
\hline
\end{tabular}}
\caption{Statistics of our dataset after preprocessing.}
\label{dataset}
\end{table}

\subsection{Experimental Setup}

\paragraph{Dataset}

Our experiments are conducted on the widely used ESConv \cite{DBLP:conf/acl/LiuZDSLYJH20}, a multi-turn conversation dataset for ES. In a conversation, the user confides personal negative situation, and the supporter provides comfort and support to improve the user's mental state. The statistics of ESConv and graph $\mathcal{G}$ after preprocessing are in Table \ref{dataset}.

\paragraph{Baselines}

(1) \textit{MoEL} \cite{DBLP:conf/emnlp/LinMSXF19}: An empathetic conversation model that uses multiple decoders to capture possible user emotions for generating.
(2) \textit{MIME} \cite{DBLP:conf/emnlp/MajumderHPLGGMP20}: An empathetic conversation model that mimics user’s emotions during responding.
(3) \textit{BlenderBot-Joint} \cite{DBLP:conf/acl/LiuZDSLYJH20}: An ESC model that prepends a predicted strategy token on the backbone of BlenderBot.
(4) \textit{MISC} \cite{DBLP:conf/acl/TuLC0W022}: An ESC model that fuses commonsense.
(5) \textit{GLHG} \cite{DBLP:conf/ijcai/00080XXSL22}: A commonsense-based ESC model that designs a global-to-local graph.
(6) We design \textit{Bart-Joint} by replacing the backbone of BlenderBot-Joint with Bart \cite{DBLP:conf/acl/LewisLGGMLSZ20}. It achieves comparable performance to \textit{MultiESC} \cite{DBLP:journals/corr/abs-2210-04242} as its replacement since MultiESC's code is unavailable.

\paragraph{Implementation Details}

We implement all models with Pytorch, and all pretrained models (i.e., BlenderBot, Bart) use small versions. We set the number of steps $K=2$ and reward weights $w_{cES}=w_{cDC}=0.1,w_{tES}=w_{fDC}=1.0$ (selected using a grid-search approach with two values \{0.1, 1.0\} for each hyperparameter). We extract $M=10$ emotional reactions for each utterance. The maximum number of conversation turn $MT$ is set to 10. The discount factor $\gamma$ is 0.99, the hyperparameter $\alpha$ is 1e-5, and the batch size is 16. We use Adam optimizer \cite{DBLP:journals/corr/KingmaB14} with an initial learning rate of 2e-5 and a linear warmup of 120 steps for training on a GPU-V100 machine. The warm start stage is trained for 5 epochs, and the joint training stage is set to 3 epochs. The decoding settings are consistent with \citet{DBLP:conf/acl/LiuZDSLYJH20}. For a fair comparison, all baselines with available codes are reproduced under the same setting.

\begin{table*}[t]
\centering
\resizebox{.95\textwidth}{!}{
\begin{tabular}{l c c c c c c c c c c c c}
\hline
    Models & PPL$\downarrow$ & B-1$\uparrow$ & B-2$\uparrow$ & B-3$\uparrow$ & D-1$\uparrow$ & D-2$\uparrow$ & D-3$\uparrow$ & \textit{cES}$\uparrow$ & \textit{tES}$\uparrow$ & \textit{cDC}$\uparrow$ & \textit{fDC}$\uparrow$ & Len \\
\hline
    MoEL & 112.34 & 18.14 & 6.77 & 3.22 & 2.43 & 17.03 & 38.08 & 0.658 & 0.390 & 0.391 & 0.384 & 20.36 \\
    MIME & 68.49 & 15.89 & 6.58 & 3.27 & 2.02 & 10.51 & 22.60 & 0.598 & 0.370 & 0.450 & 0.412 & 19.44 \\
    BlenderBot-Joint & \textbf{14.78} & 17.97 & 7.17 & 3.31 & 4.56 & 24.65 & 49.71 & 0.611 & 0.398 & 0.710 & 0.459 & 17.69 \\
    MISC & 16.16 & - & 7.31 & - & 4.41 & 19.71 & - & - & - & - & - & - \\
    GLHG & 15.67 & 19.66 & 7.57 & 3.74 & 3.50 & 21.61 & - & - & - & - & - & - \\
    Bart-Joint & 16.05 & \textbf{19.99} & \textbf{7.92} & \textbf{3.93} & 4.24 & 21.98 & 43.33 & 0.635 & 0.402 & \textbf{0.723} & \textbf{0.475} & 18.85 \\
\hline
    \textbf{\textsc{Supporter}} & 15.37 & 19.50 & 7.49 & 3.58 & \textbf{4.93} & \textbf{27.73} & \textbf{53.78} & \textbf{0.743} & \textbf{0.409} & 0.681 & 0.472 & 18.37 \\
\hline
    w/o EmoExperts & 15.35 & 18.32 & 7.12 & 3.38 & 4.79 & 27.20 & 53.01 & 0.711 & 0.392 & 0.679 & 0.460 & 18.14 \\
    w/o KwsExperts & 15.54 & 17.76 & 6.74 & 3.19 & 4.69 & 26.16 & 50.92 & 0.728 & 0.394 & 0.636 & 0.443 & 17.72 \\
    w/o Multi-Task & 15.49 & 16.79 & 6.54 & 3.18 & 4.78 & 27.17 & 53.45 & 0.651 & 0.399 & 0.651 & 0.450 & 16.48 \\
\hdashline
    w/o ESRewards & 15.46 & 18.49 & 7.10 & 3.36 & 4.69 & 26.92 & 52.49 & 0.664 & 0.391 & 0.660 & 0.457 & 18.41 \\
    w/o DCRewards & 15.43 & 17.28 & 6.80 & 3.25 & 4.80 & 27.45 & 53.04 & 0.707 & 0.401 & 0.652 & 0.448 & 17.12 \\
    w/o ExpertPolicy & 15.54 & 18.30 & 7.23 & 3.54 & 4.75 & 27.23 & 52.85 & 0.683 & 0.395 & 0.657 & 0.454 & 18.54 \\
\hdashline
    Warm-Start Only & 15.03 & 17.42 & 6.74 & 3.21 & 4.67 & 26.24 & 51.82 & 0.629 & 0.402 & 0.644 & 0.444 & 17.35 \\
    w/o Warm-Start & 15.01 & 17.98 & 6.86 & 3.18 & 4.55 & 26.06 & 51.62 & 0.673 & 0.403 & 0.638 & 0.453 & 18.26 \\
\hline
\end{tabular}}
\caption{Automatic evaluation results. ``Len'' indicates the average length of the generated responses.}
\label{overall}
\end{table*}

\subsection{Automatic Evaluation}

We adopt Perplexity (PPL), Bleu (B-\textit{n}) and Distinct (D-\textit{n}) to evaluate the general generation quality and diversity of the models. To measure how well the generated responses achieve goals, we define (1) ES scores containing conversation-level (\textit{cES}) and turn-level (\textit{tES}), i.e., $r_{cES}$ and $r_{tES}$, measure the elicitation intensity of positive emotion involving conversation progress and the perceived intensity to the user's next turn emotion; (2) Dialogue coherence scores containing contextual (\textit{cDC}) and future (\textit{fDC}), i.e., $r_{cDC}$ and $r_{fDC}$, measure the coherence with the context and the user's future utterance.

\paragraph{Overall Performance}

In Table \ref{overall}, compared with all baselines, our \textsc{Supporter} achieves the most diverse expressions and highest ES (12.9\% outperforms the second best MoEL on \textit{cES}) while maintaining competitive dialogue quality (\textit{PPL}, \textit{Bleu}) and coherence (\textit{cDC}, \textit{fDC}). Supportive responses generated by MoEL are often accompanied by low diversity and low coherence due to the retelling of generic responses (e.g., \textit{“I am glad I could help you”} with high positive emotion) that are found from its outputs. Bart-based models benefit from robust sequence modeling \cite{DBLP:conf/acl/LewisLGGMLSZ20} with inherent advantages in coherence and Bleu but perform poorly in ES and diversity. The contextual coherence (\textit{cDC}) of our \textsc{Supporter} is inferior to BlenderBot-Joint, which is acceptable as ES for positive emotion elicitation needs to sacrifice a little coherence to jump out of negative topics.

\paragraph{Ablation Study}

In Table \ref{overall}:
\textbf{First}, we remove the emotion experts (w/o EmoExperts), keyword experts (w/o KwsExperts), and the multi-task associated with the experts (w/o Multi-Task), respectively. Emotion experts mainly act on ES, including \textit{cES} and \textit{tES}. Keyword experts contribute significantly to dialogue coherence, including \textit{cDC} and \textit{fDC}. Multi-task training endows experts with specific abilities and thus has an impressive impact on overall performance.
\textbf{Second}, we remove the ES rewards (w/o ESRewards) and dialogue coherence rewards (w/o DCRewards), respectively. The former improves positive support, and the latter maintains grounded expression. Therefore, besides achieving their own goals, they also benefit dialogue diversity and quality, respectively. Moreover, we replace the expert selection policy network with random sampling (w/o ExpertPolicy). Random experts lead to uncertainty in decision-making and thus damage overall performance, especially on ES and coherence.
\textbf{Third}, we test using only warm start and without joint training (Warm-Start Only) as well as without warm start and only joint training (w/o Warm-Start). The former reaches comparable or even worse results than the baselines, and the latter greedily achieves the goal of maximizing the rewards resulting in low dialogue quality.

\begin{table*}[t]
\centering
\resizebox{.95\textwidth}{!}{
\begin{tabular}{l c c c c c c c c c c c c}
\hline
    \multirow{2}{*}{\makecell[l]{\textsc{Supporter} vs.}} & \multicolumn{3}{c}{BlenderBot-Joint} &  \multicolumn{3}{c}{Bart-Joint} & \multicolumn{3}{c}{w/o EmoExperts} & \multicolumn{3}{c}{w/o ExpertPolicy} \\
    & Win & Lose & Tie & Win & Lose & Tie & Win & Lose & Tie & Win & Lose & Tie \\
\hline
    Fluency & \textbf{67.5}$^\ddagger$ & 23.7 & 8.8 & \textbf{66.5}$^\ddagger$ & 26.5 & 7.0 & \textbf{44.5}$^\dagger$ & 40.0 & 15.5 & \textbf{42.9}$^\dagger$ & 37.5 & 19.6 \\
    Informativeness & \textbf{55.2}$^\ddagger$ & 40.7 & 4.1 & \textbf{56.7}$^\ddagger$ & 38.8 & 4.5 & \textbf{48.6}$^\ddagger$ & 36.8 & 14.6 & \textbf{38.5} & 35.9 & 25.6 \\
    Coherence & \textbf{53.8}$^\ddagger$ & 31.8 & 14.4 & \textbf{45.4} & 43.8 & 10.8 & \textbf{53.7}$^\ddagger$ & 35.7 & 10.6 & \textbf{55.1}$^\ddagger$ & 32.4 & 12.5 \\
    Supportiveness & \textbf{59.2}$^\ddagger$ & 34.1 & 6.7 & \textbf{51.4}$^\ddagger$ & 37.6 & 11.0 & \textbf{54.5}$^\ddagger$ & 33.4 & 12.1 & \textbf{51.4}$^\ddagger$ & 34.3 & 14.3 \\
\hline
    Overall & \textbf{56.5}$^\ddagger$ & 30.4 & 13.1 & \textbf{48.6}$^\ddagger$ & 37.1 & 14.3 & \textbf{50.0}$^\ddagger$ & 34.3 & 15.7 & \textbf{49.6}$^\ddagger$ & 32.1 & 18.3 \\
\hline
\end{tabular}}
\caption{Results of interactive human evaluation (\%). $\dagger/\ddagger$ denote $p$-value $<$ 0.1/0.05 (statistical significance test).}
\label{human}
\end{table*}

\subsection{Interactive Human Evaluation}

We recruited three crowdsourcing workers and exposed them to 100 negative situations randomly sampled from the test set. They were asked to engage in multi-turn conversation with the models to simulate the process of seeking ES and to choose the better one (Win) from a model pair by considering five aspects, respectively:
(1) Fluency: which bot’s response is more fluent and understandable? 
(2) Informativeness: which bot's response is more diverse and specific, and contains more information? 
(3) Coherence: which bot's response is more coherent with context in a multi-turn conversation? 
(4) Supportiveness: which bot provides more effective ES, i.e., is more likely to elicit users to change their emotions from negative to positive?
(5) Overall: generally, which bot is more preferred?

As in Table \ref{human}, from the comparison with baselines, we found that a single incoherent response (\textit{cDC} in Table \ref{overall}) has less impact on the coherence of the overall multi-turn conversation.
Comparisons with variants of \textsc{Supporter} demonstrate that key components of our model, i.e., emotion experts and expert selection policy, lead to significant advantages in the overall performance.

\subsection{Qualitative Analysis}

\begin{figure}[t]
\centering
\includegraphics[width=.7\columnwidth]{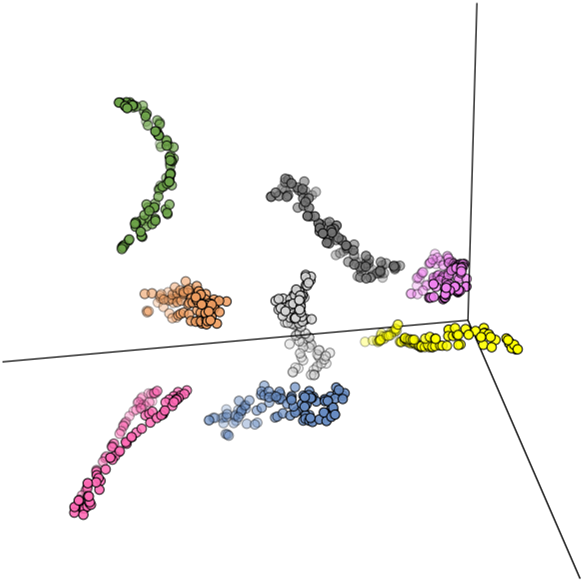} 
\caption{Latent space visualization of experts. Separate clusters show MoE has diverse and specific semantics.}
\label{expert}
\end{figure}

\paragraph{Specificity of Experts}

To analyze the quality of the experts, we show  the specificity of the experts learned by \textsc{Supporter}. As shown in Figure \ref{expert}, we visualize the latent space of experts using $t$-SNE on 200 conversation samples. The latent space distributions of multi-task-associated experts are clearly separated and clustered in specific regions. Some overlap is also intuitive due to the similarity between experts with the same polarity, e.g., contextual and future positive emotion experts. This verifies our MoE has diverse and specific semantics and the superiority of multi-task learning.

\begin{figure}[t]
\centering
\includegraphics[width=.95\columnwidth]{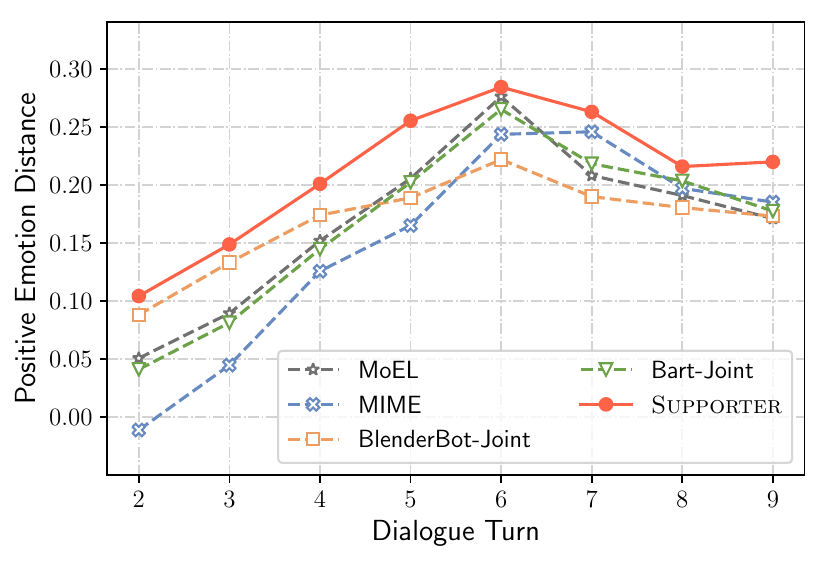} 
\caption{\textsc{Supporter} progressively enhances the elicitation intensity and exhibits robust adjustment ability in the later stage of the conversation.}
\label{ped}
\end{figure}

\paragraph{Adjustability of Elicitation}

To further explore the adjustability of elicitation intensity of positive emotion in multi-turn conversation, we analyze the trend of positive emotion distance with the dialogue turns, i.e., $PED=f_{ES}(y)-\frac{1}{T} \sum_{t=1}^T f_{ES}\left(c_t\right)$. As shown in Figure \ref{ped}, the PED score of all models tends to rise first and then fall. In the early stage of the conversation (turn$<$6), \textsc{Supporter} keeps the same trend as the empathy model (i.e., MoEL, MIME) and gradually increases the intensity of elicitation. This is attributed to our encouragement that it should progressively transform the conversation from empathy-dominated to elicitation-dominated. In the later stage of the conversation (turn$>$6), \textsc{Supporter} still maintains a higher level of elicitation than baselines and shows robust adjustment ability.

\begin{table}[t]
\centering
\resizebox{.95\columnwidth}{!}{
\begin{tabular}{l c c c c c c}
\hline
    Models & D-1 & B-2 & \textit{cES} & \textit{tES} & \textit{cDC} & \textit{fDC} \\
\hline
    \textsc{Supporter}$_{K=1}$ & 4.40 & 7.55 & 0.801 & 0.382 & 0.668 & 0.466 \\
    \textsc{Supporter}$_{K=2}$ & 4.93 & 7.49 & 0.743 & 0.409 & 0.681 & 0.472 \\
    \textsc{Supporter}$_{K=3}$ & 5.22 & 6.71 & 0.699 & 0.405 & 0.657 & 0.459 \\
    \textsc{Supporter}$_{K=4}$ & 5.05 & 6.10 & 0.673 & 0.413 & 0.594 & 0.431 \\
\hline
\end{tabular}}
\caption{Parameter analysis for iteration steps $K$. \textsc{Supporter} outperforms the best baselines in most settings.}
\label{actionnum}
\end{table}

\subsection{Parameter Analysis} 
\label{sec:parameter}
We further analyze the impact of the number of iteration steps $K$. In Table \ref{actionnum}, with the increase of steps, diversity and \textit{tES} show an upward trend, while other metrics show a downward one. This happens possibly because the informativeness of the generated responses increases with selected experts, making it possible to lose focus and thus lead to poor dialogue quality. Furthermore, \textsc{Supporter} outperforms the best baselines in most cases, confirming its effectiveness.

\section{Conclusions}

In this paper, we introduce a new paradigm to formalize multi-turn ESC as a process of positive emotion elicitation and propose an MoE-based reinforcement learning model \textsc{Supporter} with well-designed ES and dialogue coherence rewards. Extensive experiments verify the superiority of our model in providing effective ES for positive emotion elicitation while maintaining conversational goals including coherence. Our work will facilitate future work to develop ESC with positive emotion elicitation for improving the users’ mental state.

\section*{Limitations}

We discuss three limitations of this work as follows.

The first one is the instability of reinforcement learning. Reward-driven policy learning is an essential advantage of this work because it is better equipped with the positive emotion-driven process of ESC than existing works and can model flexible ESC expression beyond the training data. However, this flexibility also suffers from instability, which calls for additional knowledge or strategies to refine the learning process.

The second one is the need for further reference to psychological theory. An advantage of our work is to learn posterior ESC patterns integrating the dialogue context and future feedback in the form of rewards. However, there is still other valuable prior knowledge to be referred from psychology studies, e.g., the CBT (cognitive-behavioral therapy) methods. This kind of prior knowledge can be used as additional knowledge to refine the learning process as mentioned in the first limitation.

The third one is that the reward design can be further optimized. The ideal case is to construct a high-quality dataset with human-feedback labels for training reward model (e.g., the constructed example of ChatGPT). At the same time, the larger parameter of the reward model, the more conducive it is to learn a robust policy and avoid it overfitting to the reward function. However, such optimizations need a trade-off with cost.

\section*{Ethical Considerations}

In this paper, the ESConv dataset used in our experiments is a publicly-available benchmark for emotional support conversation, which does not contain sensitive and personal information as well as unethical language. Our work builds on this dataset to study positive emotion elicitation to improve the user's mental state. Therefore, we focus on constructing a dialogue system to provide emotional support from families and friends in the daily scenarios limited by this dataset rather than professional psychological counseling or psychological treatment. For risky non-daily scenarios such as self-harm or suicide-related conversations, we do not claim that the dialogue system we built has a treatment or improvement effect on them. Additionally, we also ensure the anonymity of our interactive human evaluation. We believe our work meets ACL's Code of Ethics.

\section*{Acknowledgements}

This work was supported by the National Science Foundation for Distinguished Young Scholars (with No. 62125604). This work was also supported by the Guoqiang Institute of Tsinghua University, with Grant No. 2020GQG0005. This work was also supported by Tsinghua Precision Medicine Foundation.

This work was also supported by the National Natural Science Foundation of China (with No. 62272340, 61876128, 62276187).

\bibliography{anthology,ESC}
\bibliographystyle{acl_natbib}

\clearpage

\appendix

\begin{figure}[t]
\centering
\includegraphics[width=1.\columnwidth]{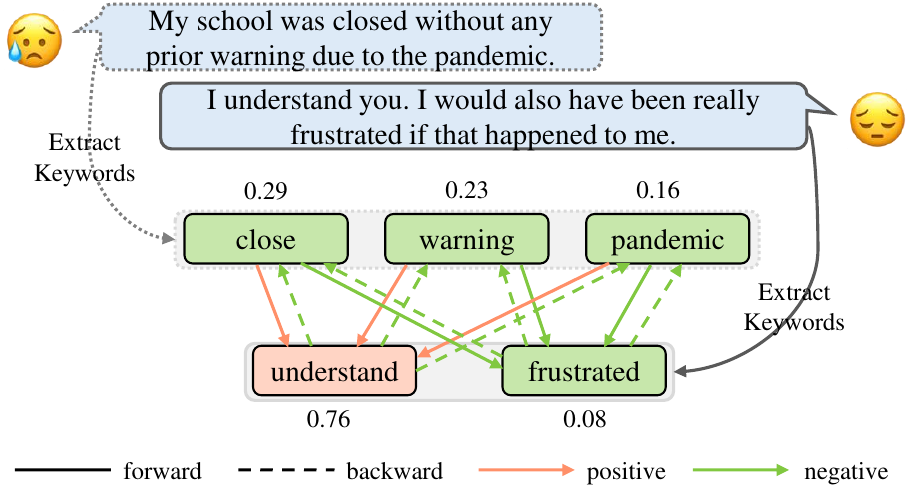} 
\caption{A construction example of the bidirectional emotion keyword graph $\mathcal{G}$. The valence (i.e., the number in the figure) is used to identify \colorbox{positive}{\textit{positive}} and \colorbox{negative}{\textit{negative}} keywords.}
\label{graph}
\end{figure}

\section{Bidirectional Emotion Keyword Graph}
\label{sec:graph}

A construction example of the bidirectional emotion keyword graph $\mathcal{G}$ is in Figure \ref{graph}.

\paragraph{One-hop Reasoning on Graph $\mathcal{G}$}
For the contextual keyword “\textit{close}”, its one-hop neighbor reasoned by the “\textit{forward-positive}” relation is “\textit{understand}”, and the one reasoned by the “\textit{forward-negative}” relation is “\textit{frustrated}”. Further, the one-hop neighbors reasoned by the “\textit{forward}” relation are the union of the one-hop neighbors of the above two relations, i.e., “\textit{understand}” and “\textit{frustrated}”.
For the keyword “\textit{frustrated}” of the response, it cannot reason the one-hop neighbor using the “\textit{backward-positive}” relation. Therefore, its one-hop neighbors reasoned by the “\textit{backward}” relation are the same as the one-hop neighbors reasoned by the “\textit{backward-negative}” relation, i.e., “\textit{close}”, “\textit{warning}”, and “\textit{pandemic}”.

\paragraph{Multi-hop Reasoning on Graph $\mathcal{G}$}
Taking the “\textit{forward} $\rightarrow$ \textit{forward} $\rightarrow$ \textit{backward-positive}” multi-hop reasoning as an example, using the “\textit{forward}” relationship for the contextual keywords to perform one-hop reasoning can obtain the set of neighbors that contain the keywords of the response, which we regard as the extended keyword set of the response determined by the context. 
Using the keywords in this set as a starting point to perform the second-hop reasoning by the “\textit{forward}” relation can result in the expanded keyword set of the user’s future utterance (i.e., the user’s next turn utterance) determined by the response. 
Further, similarly, the third-hop reasoning is performed using the “\textit{backward-positive}” relation to determine the extended positive keywords set of the response coherent to the future utterance.

\section{Case Study}
\label{sec:casestudy}

In Table \ref{case}, two cases in multi-turn conversation generated by five models are selected. 
We found that the advantage of \textsc{Supporter} is that it is more likely to elicit users' emotions to transit positively by adaptively adjusting the intensity of empathy and elicitation. For example, low turns have a tone of strong empathy (e.g., “\textit{that is so awful}”, “\textit{you are struggling financially and that you have not been able to afford your kids presents}”). As the conversation deepens, the elicitation intensity of positive emotion progressively increases (e.g., “\textit{do you know anyone who has gone through similar situations?}”, “\textit{it is okay to be proud of yourself for being able to make them}”), and finally the expression with strong elicitation will help users jump out of negative situations (e.g., “\textit{I am sure you will be successful in the new job market}”, “\textit{you can really enjoy seeing them happy}”). During these multi-turn conversations, \textsc{Supporter} effectively provides ES with diverse expressions while maintaining the dialogue coherence, being consistent with the observations from automatic and interactive human evaluations.

\begin{table*}[t]
\centering
\resizebox{\linewidth}{!}{
\begin{tabular}{l l}
\hline
\hline
    \multirow{2}{*}{\makecell[l]{Situation\#1}} & I have been laid out for not performing. But the actual situation is the customer was satisfied my service. But the expectation of \\
    & the management was not to do that. \\
\hline
    \multirow{3}{*}{\makecell[l]{Context \\ (1$^{st}$-2$^{nd}$ turn)}} 
    & \textit{User}: Hi! \\
    & \textit{Bot}: Hi can I help you today? I am here to listen. Do you have anything happen recently that has you feeling badly? \\
    & \textit{User}: I am laid off. \\
\hline
    MoEL & I am sorry to hear that. What is going on with your thoughts? \\
    MIME & I am sorry to hear that. I am sorry to hear that. I am so sorry to hear that. \\
    BlenderBot-Joint & I am sorry to hear that. Is there something you would like to talk about? \\
    Bart-Joint & I have been laid off before. I was laid off too. \\
\hdashline
    \textbf{\textsc{Supporter}} & \textbf{Oh no, that is so awful. I am so sorry to hear that. I know that feeling well, especially when you have this situation.} \\
\hdashline
    Ground-Truth & Oh I am really sorry to hear that, did you have the same job for a long time? That sounds very difficult to deal with. \\
\hline
\multirow{5}{*}{\makecell[l]{Context \\ (3$^{rd}$-5$^{th}$ turn)}} 
    & \textit{User}: I was attending a customer. He was having a grievance which I sorted out. But the management did not like that. For the past \\
    & 10 years I have been with this job. \\
    & /*......*/ \\
    & \textit{Bot}: That is really unfair and hard to deal with are you close to any family? \\
    & \textit{User}: I am not in any close to any family related to job. \\
\hline
    MoEL & I think that it is great you have a great job! \\
    MIME & It sounds like you are trying to get your job that you are not alone. \\
    BlenderBot-Joint & I can tell you are really feeling badly about this. \\
    Bart-Joint & I know that is hard to deal with. I am glad that you are reaching out to me to help out! \\
\hdashline
    \textbf{\textsc{Supporter}} & \textbf{That is a really difficult position to be in. do you know anyone who has gone through similar situations?} \\
\hdashline
    Ground-Truth & Do you have any close friends to talk to about any new job prospects? \\
\hline
\multirow{6}{*}{\makecell[l]{Context \\ (6$^{th}$-9$^{th}$ turn)}} 
    & \textit{User}: I have few friends, I have been talking to them. They also tell I was not any wrong. For the experience and qualification, I will \\
    & be getting a better job. \\
    & /*......*/ \\
    & \textit{Bot}: Oh, wow then you do have a lot of contacts and some support or at least understanding. Music is a great way to relax and that \\
    & is very positive in your life and current situation. \\
    & \textit{User}: I am so confident I will be having a better job within the next few months. \\
\hline
    \multirow{2}{*}{\makecell[l]{MoEL}} & I have found that online too. I do not know if I have heard of my work. I found out online that I could talk to you. I would suggest \\
    & looking for a new job.  \\
    MIME & I am glad to hear that! I bet you could try to get some breathing techniques. \\
    \multirow{2}{*}{\makecell[l]{BlenderBot-Joint}} & It is very difficult to think about it right now but I would definitely encourage you to take some deep breaths and take some time to \\
    & reflect on your progress. \\
    Bart-Joint & I am so happy that you are feeling better and have found some new ways to relax. \\
\hdashline
    \multirow{2}{*}{\makecell[l]{\textbf{\textsc{Supporter}}}} & \textbf{I am glad that I could be of some help to you. It sounds like you are a very strong and intelligent person. I am sure you will be} \\
    & \textbf{successful in the new job market.} \\
\hdashline
    Ground-Truth & That is a great outlook. Thank you for sharing today. \\
\hline
\hline
    Situation\#2 & I don't have enough money to buy my kids Christmas gifts.\\
\hline
\multirow{5}{*}{\makecell[l]{Context \\ (1$^{st}$-3$^{rd}$ turn)}} 
    & \textit{User}: Hi! \\
    & \textit{Bot}: How are things going? \\
    & \textit{User}: Not good. I am depressed. \\
    & \textit{Bot}: Is there anything in particular that has been bothering you recently? \\
    & \textit{User}: I do not have enough money to buy my 5 kids X-mas gifts. \\
\hline
    MoEL & I am sorry to hear that. I am sorry you are having this right now.  \\
    MIME & I understand how you are feeling. I know how hard it must be. \\
    BlenderBot-Joint & So you are feeling depressed because of covid and not enough money to buy gifts? \\
    Bart-Joint & I am sorry to hear that. I have had to give up my job to help my kids during the holidays. \\
\hdashline
    \textbf{\textsc{Supporter}} & \textbf{It seems to me that you are struggling financially and that you have not been able to afford your kids presents.} \\
\hdashline
    Ground-Truth & I am really sorry to hear that, but I am sure your love for your kids can still shine through. \\
\hline
\multirow{3}{*}{\makecell[l]{Context \\ (4$^{th}$-5$^{th}$ turn)}} 
    & \textit{User}: I do love my kids, but they are expecting some toys and stuff from Santa. \\
    & \textit{Bot}: Would it be possible to look at some more home made stuff for this year that will still be fun and meaningful? \\
    & \textit{User}: I can not craft. If I made them something it would be so bad. \\
\hline
    MoEL & I am so sorry to hear that. I am sorry you are having to go through this.  \\
    MIME & I understand how you feel. I have been there too. I know how hard it is to the next year. I am sure there are some options out there. \\
    BlenderBot-Joint & Have you considered doing crafts with your kids? \\
    \multirow{2}{*}{\makecell[l]{Bart-Joint}} & Have you considered getting some more craft supplies? Maybe something that you can use to make a list of things you would like \\
    & to get your kids for Christmas? \\
\hdashline
    \multirow{2}{*}{\makecell[l]{\textbf{\textsc{Supporter}}}} & \textbf{I know how hard it can be to make things. I know it is a terrible feeling to feel bad about yourself, but it is okay to be proud of} \\
    & \textbf{yourself for being able to make them.} \\
\hdashline
    \multirow{2}{*}{\makecell[l]{Ground-Truth}} & I know it can be really hard to feel like you can not make anything worthwhile, but I am sure there's something you could come up \\
    & with that could still put a smile on your kids' faces. \\
\hline
\multirow{4}{*}{\makecell[l]{Context \\ (6$^{th}$-7$^{th}$ turn)}} 
    & \textit{User}: Do you have any craft ideas? \\
    & \textit{Bot}: Maybe you could look at involving your kids in the process. The feeling of doing something together as a family can make it \\
    & way more fun. It could be a shared experience your family can look back on fondly later on. \\
    & \textit{User}: That is a good idea. I would like to have it done before Christmas. How about some wooden toys or a cardboard box fort? \\
\hline
    MoEL & I am so glad to hear that. I hope that you can do something fun!  \\
    MIME & I think you can do it! I have been through this, and you are not alone? \\
    \multirow{2}{*}{\makecell[l]{BlenderBot-Joint}} & That sounds like a great idea! Maybe even something for them to look forward to after Christmas. I know I love the idea of making \\
    & toys and toys! \\
    \multirow{2}{*}{\makecell[l]{Bart-Joint}} & Yes, I think you should try to get it done before Christmas, but maybe even after that. You can see how hard it is to find a good time to \\
    & do something. \\
\hdashline
    \multirow{2}{*}{\makecell[l]{\textbf{\textsc{Supporter}}}} & \textbf{I think that would be a great idea! That way you can feel like your kids are being cared for and you can really enjoy seeing} \\
    & \textbf{them happy.} \\
\hdashline
    \multirow{2}{*}{\makecell[l]{Ground-Truth}} & I have seen some people make small ornaments filled with candies and treats and other small goodies. So even if all of the gifts feel \\
    & small, they can feel like they are getting a lot of them. And you could even get them to have fun decorating and painting the ornaments! \\
\hline
\hline
\end{tabular}}
\caption{Cases generated from baselines and \textsc{Supporter}. /*......*/ indicates that some turns of dialogue are omitted.}
\label{case}
\end{table*}

\end{document}